\documentclass[sigconf, nonacm]{acmart}

\usepackage{amsmath}
\usepackage{amsthm}
\usepackage{multicol}
\usepackage{comment}
\usepackage{array}
\usepackage{eqparbox}
\usepackage{color}
\usepackage{subcaption}
\usepackage{caption} 
\usepackage{xspace}
\usepackage[framemethod=tikz]{mdframed}
\usepackage{listings}
\usepackage{paralist}
\usepackage[linesnumbered,ruled,vlined]{algorithm2e}

\mdfsetup{%
  font=\small,
  innertopmargin=2pt,
  innerbottommargin=2pt,
  innerleftmargin=8pt,
  innerrightmargin=8pt,
  frametitleaboveskip=2pt,
  frametitlebelowskip=1pt,
  frametitlealignment=\center
}
\newenvironment{prompt}[1]
{
  \vspace{1ex}
  \begin{mdframed}[frametitle={#1}]
}
{
  \end{mdframed}
}

\newcommand{\myparagraph}[1]{\vspace{1ex}\noindent\underline{\it #1.}\xspace}

\newcommand{\str}[1]{\textcolor{purple}{#1}}

\begin{document}

\title{Large Language Models as Data Preprocessors}

\newcommand{\osaka}{$^{1}$}
\newcommand{\nec}{$^{2}$}
\newcommand{\nagoya}{$^{3}$}
\newcommand{\osanag}{$^{1,3}$}

\author{\osaka Haochen Zhang, \nec Yuyang Dong, \osanag Chuan Xiao, \nec Masafumi Oyamada}

\affiliation{\country{}\osaka Osaka University, \nec NEC Corporation, \nagoya Nagoya University}

\email{{chou.koushin, chuanx}@ist.osaka-u.ac.jp, {dongyuyang, oyamada}@nec.com}

\begin{abstract}
  Large Language Models (LLMs), typified by OpenAI's GPT, have marked a significant advancement in artificial intelligence. Trained on vast amounts of text data, LLMs are capable of understanding and generating human-like text across a diverse range of topics. This study expands on the applications of LLMs, exploring their potential in data preprocessing, a critical stage in data mining and analytics applications. Aiming at tabular data, we delve into the applicability of state-of-the-art LLMs such as GPT-4 and GPT-4o for a series of preprocessing tasks, including error detection, data imputation, schema matching, and entity matching. Alongside showcasing the inherent capabilities of LLMs, we highlight their limitations, particularly in terms of computational expense and inefficiency. We propose an LLM-based framework for data preprocessing, which integrates cutting-edge prompt engineering techniques, coupled with traditional methods like contextualization and feature selection, to improve the performance and efficiency of these models. The effectiveness of LLMs in data preprocessing is evaluated through an experimental study spanning a variety of public datasets. GPT-4 emerged as a standout, achieving 100\% accuracy or F1 score on 4 of these datasets, suggesting LLMs' immense potential in these tasks. Despite certain limitations, our study underscores the promise of LLMs in this domain and anticipates future developments to overcome current hurdles.
\end{abstract}





\maketitle

\section{Introduction}
\label{sec:intro}

Large Language Models (LLMs), such as OpenAI's GPT and Meta's LLaMA, are becoming an increasingly important aspect of the AI landscape. These models, essentially ML systems, are trained on vast amounts of text data and characterized by an augmented number of parameters. They are capable of understanding and generating text across a diverse range of topics, thereby finding applications in numerous tasks. Consequently, research involving LLMs has garnered significant attention from both academia and industry. Recent endeavors have successfully leveraged LLMs for data management and mining. For instance, LLMs have been used for SQL generation~\cite{trummer2022codexdb}, database 
diagnosis~\cite{dbgpt}, data wrangling~\cite{narayan2022can}, and data analytics~\cite{cheng2023gpt}.

This paper investigates the potential of utilizing state-of-the-art (SOTA) LLMs for 
data preprocessing, a crucial step that refines data before it can be harnessed for 
downstream data mining and analytics applications. Given their comprehensive 
understanding of language semantics and structures, LLMs can identify errors or 
matches in text data. For example, they are capable of detecting spelling mistakes, 
grammar issues, contextual discrepancies, and near-duplicate records. Consequently, 
the application of LLMs in data preprocessing can pave the way for tackling tasks 
such as error detection, data imputation, schema matching, and entity matching.

While LLMs hold considerable potential for data preprocessing tasks, it is critical 
to comprehend their capabilities and limitations for effective application. Thus, 
as a preliminary study on employing LLMs for data preprocessing, this paper provides 
the following contributions.

\noindent{} (1) We examine the inherent knowledge and superior reasoning and learning abilities of LLMs, which can be further enhanced through zero- and few-shot prompting. These strengths position LLMs as competitive candidates for various data processing tasks. However, their computational expense and potential inefficiencies present challenges. We provide an analysis of these strengths and limitations in the context of data preprocessing.

\noindent{} (2) We propose a framework for LLM-based data preprocessing. This framework integrates a series of SOTA prompt engineering techniques, including zero-shot instructions, few-shot examples, batch prompting, as well as traditional approaches such as contextualization and feature selection. We specifically instruct LLMs to follow an answer format and reason before providing an answer to enhance performance. Few-shot examples are used to condition LLMs so that they can learn error criteria, means of imputation, matching conditions, etc. Batch prompting amalgamates multiple data instances in a prompt to reduce token and time costs.

\noindent{} (3) We conduct experiments on 12 datasets for four data preprocessing tasks. We evaluate popular LLMs such as GPT-3.5, GPT-4, and GPT-4o. The results indicate that GPT-4 generally outperforms existing solutions, achieving 100\% accuracy or F1 score on 4 out of 12 datasets. GPT-3.5 also delivers competitive performance and is recommended for data preprocessing. GPT-4o delivers inconsistent performance: competitive on data imputation and entity matching but mediocre on error detection and schema matching. The evaluation also sheds light on the effects of the proposed components of the solution framework on accuracy and efficiency.

\section{Preliminaries}
\label{sec:prelim}

\subsection{Data Preprocessing}
\label{sec:problem}
In this initial exploration of large language models 
(LLMs) for data preprocessing, we concentrate on tabular data. 
We target the following tasks: error detection (ED), data imputation (DI), schema matching (SM), and entity matching (EM). Other typical data preprocessing tasks, such as data fusion and data 
wrangling, are reserved for future work. 
Diverging from the traditional definition that presents the entire dataset and finds or fixes all the errors (or matches, etc.) within, we define the problem by handling one record (or a pair) at a time, so the prompt to an LLM can be easily written. We term each input object a \emph{data instance}, i.e., a tuple for ED and DI, a pair of attributes for SM and a pair of tuples for EM.

\subsection{Large Language Models}
\label{sec:llm}
LLMs have become one of the hottest topics in 
the AI research community~\cite{zhao2023survey}. 
We discuss the strengths and 
limitations of using LLMs for data preprocessing.

\myparagraph{Strengths}
\begin{inparaenum} [(1)]
  \item 
  With their comprehensive understanding of 
  language semantics and structures, and the knowledge acquired through training 
  on vast amounts of text data, LLMs are general problem solvers capable of 
  identifying errors, anomalies, and matches in textual data, without needing 
  human-engineered rules~\cite{razniewski2021language} or fine-tuning for specific 
  tasks.
  \item Most LLMs provide a prompting interface with which users can interact and 
  assign tasks in natural language, contrasting with existing data preprocessing 
  solutions that require computer programming or specific tools (e.g., 
  HoloClean~\cite{rekatsinas2017holoclean} and Magellan~\cite{konda2016magellan}).
  \item LLMs are excellent reasoners~\cite{kojima2022large}, enabling them to not 
  only return data preprocessing results but also provide the reasons for these 
  results. In this sense, their answers are more interpretable than those of other 
  DL approaches.
  \item LLMs can be conditioned by few-shot prompting~\cite{brown2020language}. As 
  such, we can tune the criteria for data preprocessing tasks (e.g., the 
  degree of matching) using few-shot examples. 
\end{inparaenum}

\myparagraph{Limitations}
\begin{inparaenum} [(1)]
  \item For data preprocessing, one of the major limitations is the difficulty in domain specification~\cite{narayan2022can}. When dealing with data from highly specialized 
  domains, training LLMs can be costly and sometimes even impossible due to frozen 
  parameters.
  \item LLMs sometimes generate text that is plausible-sounding but factually incorrect or 
  nonsensical, as they lack a fundamental understanding of the world and rely solely on 
  the patterns they learned during training.
  \item LLMs often require substantial computational resources, thereby increasing the 
  cost of use and compromising the efficiency and scalability of data preprocessing on 
  large-scale data.
\end{inparaenum}
\section{Method}
\label{sec:method}
\begin{figure} [!t]
    \centering
    \includegraphics[width=\linewidth]{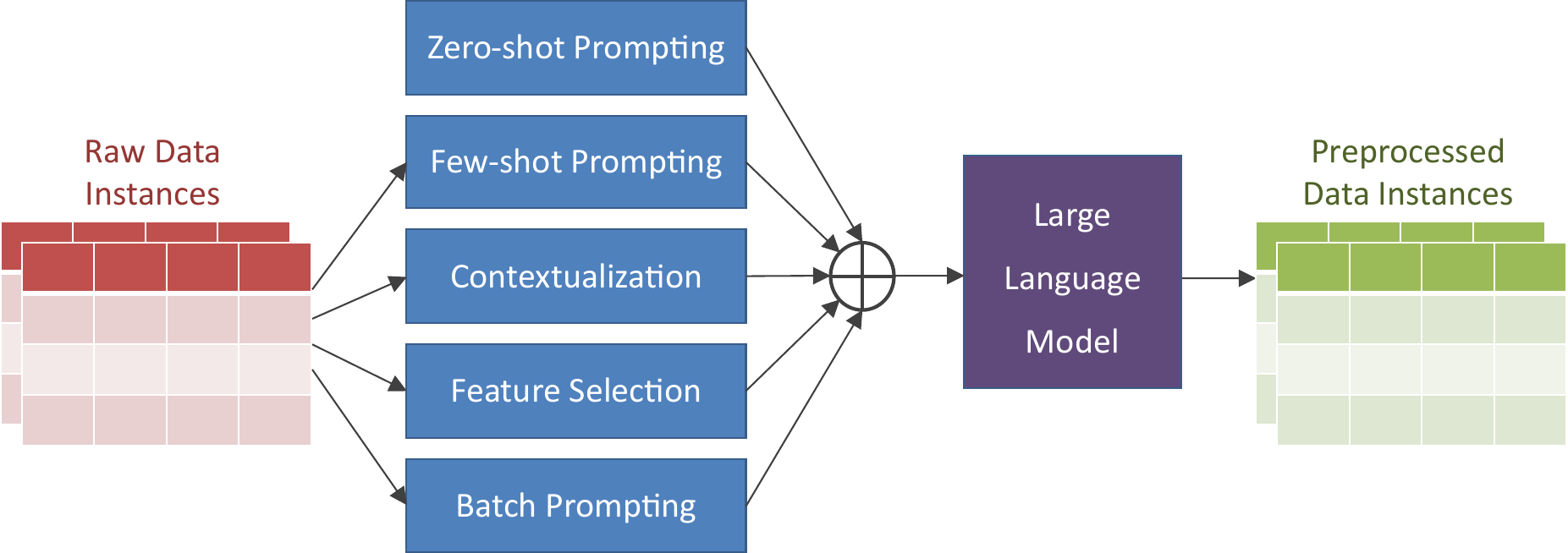}
    \caption{Framework of data preprocessing with an LLM.}
    \label{fig:framework}
\end{figure}

Figure~\ref{fig:framework} illustrates our data preprocessing framework, which consists of several modules that construct the prompt serving as the input to the LLM. 
We design a prompt template as 
follows.
\begin{prompt}{}
  \str{You are a database engineer.}\\
  \str{[Zero-shot prompt]}\\
  \str{[Few-shot prompt]}\\
  \str{[Batch prompt]}
\end{prompt}
Initially, we instruct the LLM to impersonate a database engineer. Other prompt 
components are marked within \str{[]} and will be discussed throughout this section.

\subsection{Zero-shot Prompting}
\label{sec:zero-shot}
Zero-shot prompting is a technique that guides LLMs to generate the desired 
output. It has been demonstrated to effectively enhance the reasoning abilities 
of LLMs~\cite{kojima2022large}. We employ zero-shot prompting to specify both 
the task and the answer format. Specifically, we adhere to the chain-of-thought 
paradigm~\cite{wei2022chain}, in which the LLM is expected to reason before 
delivering the answer. An example of a zero-shot prompt for DI is as follows:
\begin{prompt}{}
  \str{You are requested to infer the value of the "city" attribute based on the 
  values of other attributes.\\
  MUST answer each question in two lines. 
  In the first line, you give the reason for the inference. In the second 
  line, you ONLY give the value of the "city" attribute.}
\end{prompt}

We design specific zero-shot prompts for ED and DI. For ED, since 
we provide the entire record $r$ but ask the LLM to detect an error in one 
attribute $r_j$ at a time, the LLM might erroneously identify an error in 
attribute $r_{j'}$, where $j' \neq j$. To avoid this, we prompt the LLM to 
confirm the target attribute with: \str{Please confirm 
the target attribute in your reason for inference}. For DI, we provide a hint 
about the data type of the attribute to be imputed. For example, given the hint 
\str{The "hoursperweek" attribute can be a range of integers}, the LLM will 
respond with a range instead of a single number.

\subsection{Few-shot Prompting}
\label{sec:few-shot}
Few-shot prompting~\cite{brown2020language} involves providing a small selection 
of examples to condition the LLM for tasks that deviate from its pre-training 
objectives (e.g., text completion and code generation). We apply few-shot prompting 
by manually selecting a subset of data instances from the dataset and labeling 
them. For example, the few-shot examples for DI are presented as follows:
\begin{prompt}{}
  \str{Users:}\\
  \str{Question 1:} 
  \str{Record is [Data Instance 1]. What is the city?}\\
  \str{...}\\
  \str{Assistant:}\\
  \str{Answer 1:} 
  \str{[Reason 1]}\\
  \str{[Answer 1]}\\
  \str{...}
\end{prompt}
The data instances here adhere to the contextualization introduced in 
Section~\ref{sec:contextualization}. Users are required to provide plausible 
reasoning for few-shot examples. For instance, given \str{[name: "carey's 
corner", addr: "1215 powers ferry rd.", phone: "770-933-0909", type: 
"hamburgers", city: ???]} as \str{[Data Instance 1]}, \str{[Reason 1]} would 
be \str{The phone number "770" suggests that the city should be either Atlanta 
or Marietta in Georgia. The addr attribute suggests a place in Marietta.}, and 
\str{[Answer 1]} would be \str{Marietta}.

\subsection{Contextualization}
\label{sec:contextualization}
Given that LLMs intake raw text as input, we convert the contents in each data 
instance to a text sequence in the following format:
\begin{prompt}{}
    \str{[$x_1$.name: "$x_1$.value", $\ldots$, $x_n$.name: "$x_n$.value"]}
\end{prompt}
$x_i$ denotes the $i$-th attribute of a data instance, \texttt{name} denotes 
to the attribute name, \texttt{value} denotes the cell value, and $n$ is 
the number of input attributes. Specifically, we use \str{???} to denote missing 
values for DI, and $x_1$.name = \str{name} and $x_2$.name = \str{description} for SM.

\begin{table*}[!t]
  \small
  \centering
  \caption{Comparison with baselines, measured in accuracy (\%) for data imputation and F1 score (\%) 
  for the other tasks. LLMs are equipped with the best setting. ``N/A'' denotes not applicable 
  or not reported in their original papers.}
  \resizebox{\linewidth}{!}{%
  \begin{tabular}{l|p{.05\linewidth}<{\centering}p{.07\linewidth}<{\centering}|p{.05\linewidth}<{\centering}p{.09\linewidth}<{\centering}|p{.09\linewidth}<{\centering}|p{.05\linewidth}<{\centering}p{.04\linewidth}<{\centering}p{.04\linewidth}<{\centering}p{.04\linewidth}<{\centering}p{.05\linewidth}<{\centering}p{.06\linewidth}<{\centering}p{.06\linewidth}<{\centering}} \hline
    & \multicolumn{2}{c|}{Error Detection} & \multicolumn{2}{c|}{Data Imputation} & Schema Matching & \multicolumn{7}{c}{Entity Matching} \\ \hline
    Methods     & Adult & Hospital & Buy & Restaurant & Synthea & Amazon-Google & Beer & DBLP-ACM & DBLP-Google & Fodors-Zagats & iTunes-Amazon & Walmart-Amazon \\ \hline 
    HoloClean   & 54.5 & 51.4 & N/A  & N/A  & N/A  & N/A  & N/A  & N/A  & N/A  & N/A  & N/A  & N/A \\
    HoloDetect  & 99.1 & 94.4 & N/A  & N/A  & N/A  & N/A  & N/A  & N/A  & N/A  & N/A  & N/A  & N/A \\
    IPM         & N/A  & N/A  & 96.5 & 77.2 & N/A  & N/A  & N/A  & N/A  & N/A  & N/A  & N/A  & N/A \\
    SMAT        & N/A  & N/A  & N/A  & N/A  & 38.5 & N/A  & N/A  & N/A  & N/A  & N/A  & N/A  & N/A \\
    Magellan    & N/A  & N/A  & N/A  & N/A  & N/A  & 49.1 & 78.8 & 98.4 & 92.3 & \textbf{100} & 91.2 & 71.9 \\ 
    Ditto       & N/A  & N/A  & N/A  & N/A  & N/A  & \textbf{75.6} & 94.4 & \textbf{99.0} & 95.6 & \textbf{100} & 97.1 & 86.8 \\
    Unicorn & N/A & N/A & N/A & N/A & N/A & N/A & 90.3 & N/A & 95.6 & \textbf{100} & 96.4 & 86.9 \\
    Unicorn ++ & N/A & N/A & N/A & N/A & N/A & N/A & 87.5 & N/A & \textbf{96.2} & 97.7 & 98.2 & 86.9 \\
    Table-GPT & N/A & N/A & N/A & N/A & N/A & 70.1 & 96.3 & 93.8 & 92.4 & 97.7 & 92.9 & 82.4 \\
    GPT-3       & \textbf{99.1} & \textbf{97.8} & 98.5 & 88.4 & 45.2 & 63.5 & \textbf{100}  & 96.6 & 83.8 & \textbf{100} & 98.2 & 87.0 \\
    GPT-3.5     & 92.0 & 90.7 & 98.5 & 94.2 & 57.1 & 66.5 & 96.3 & 94.9 & 76.1 & \textbf{100} & 96.4 & 86.2 \\
    GPT-4       & 92.0 & 90.7 & \textbf{100} & \textbf{97.7} & \textbf{66.7} & 74.2 & \textbf{100} & 97.4 & 91.9 & \textbf{100} & \textbf{100} & \textbf{90.3} \\
    GPT-4o      & 83.6 & 44.8 & \textbf{100} & 90.7 & 6.6 & 70.9 & 90.3 & 95.9 & 90.4 & 93.6 & 98.2 & 79.2 \\ 
    \hline
  \end{tabular}
  }
  \label{tab:exp-comparison-baselines}
\end{table*}

\begin{table*}[!t]
  \small
  \centering
  \caption{Ablation study, measured in accuracy (\%) for data imputation and F1 score (\%) 
  for the other tasks, using GPT-3.5. \textsf{ZS-T} denotes zero-shot task specification. 
  \textsf{FS} denotes few-shots. \textsf{B} denotes batch prompting. \textsf{ZS-R} 
  denotes zero-shot reasoning.}
  \resizebox{\linewidth}{!}{%
  \begin{tabular}{l|p{.05\linewidth}<{\centering}p{.07\linewidth}<{\centering}|p{.05\linewidth}<{\centering}p{.09\linewidth}<{\centering}|p{.09\linewidth}<{\centering}|p{.05\linewidth}<{\centering}p{.04\linewidth}<{\centering}p{.04\linewidth}<{\centering}p{.04\linewidth}<{\centering}p{.05\linewidth}<{\centering}p{.06\linewidth}<{\centering}p{.06\linewidth}<{\centering}} \hline
    & \multicolumn{2}{c|}{Error Detection} & \multicolumn{2}{c|}{Data Imputation} & Schema Matching & \multicolumn{7}{c}{Entity Matching} \\ \hline
    Components     & Adult & Hospital & Buy & Restaurant & Synthea & Amazon-Google & Beer & DBLP-ACM & DBLP-Google & Fodors-Zagats & iTunes-Amazon & Walmart-Amazon \\ \hline 
    \textsf{ZS-T} & 25.9 & 18.4 & 86.2 & 81.4 & 18.2 & 54.7 & 83.3 & 94.7 & 58.5 & 92.7 & 80.0 & 81.5 \\
    \textsf{ZS-T}+\textsf{B} & 37.8 & 19.1 & 83.1 & 81.4 & 17.4 & 60.1 & 78.3 & 94.9 & 59.6 & 92.7 & 83.9 & 81.6 \\
    \textsf{ZS-T}+\textsf{B}+\textsf{ZS-R} & 46.3 & 26.2 & 89.2 & 65.1 & 5.9 & 45.8 & 50.0 & 72.6 & 47.6 & 92.7 & 82.0 & 60.7 \\
    \textsf{ZS-T}+\textsf{FS} & 59.3 & 59.4 & 96.9 & 90.7 & 57.1 & 66.3 & \textbf{96.3} & \textbf{97.0} & 74.6 & \textbf{100} & \textbf{96.4} & 85.6 \\
    \textsf{ZS-T}+\textsf{FS}+\textsf{B} & 58.1 & 56.1 & 96.9 & 86.2 & 53.3 & \textbf{66.5} & \textbf{96.3} & 96.2 & \textbf{76.1} & 97.8 & 94.7 & \textbf{86.2} \\
    \textsf{ZS-T}+\textsf{FS}+\textsf{B}+\textsf{ZS-R} & \textbf{92.0} & \textbf{90.7} & \textbf{98.5} & \textbf{94.2} & \textbf{61.5} & 60.1 & 92.3 & 95.7 & 60.0 & 97.8 & \textbf{96.4} & 84.0 \\ \hline
  \end{tabular}
  }
  \label{tab:exp-ablation-study}
\end{table*}

\subsection{Feature Selection}
If metadata is available, users can manually select a subset of features to improve 
performance. For instance, when imputing a restaurant's location, the phone number 
and street name are relevant features, while the restaurant's name and type (Asian, 
Italian, etc.) are irrelevant. Therefore, users may choose to use only the phone 
number and street name as attributes in the above prompt.

\subsection{Batch Prompting}
Considering the significant token and time cost of LLMs, batch 
prompting~\cite{cheng2023batch} was proposed to enable the LLM to run inference in 
batches, rather than processing one sample at a time. To implement this technique, 
multiple data instances are presented in a single prompt, and the LLM is instructed 
to answer all of them. For example, for DI, the prompt is the same as the first part 
of few-shot prompting (i.e., the part before \str{Assistant:}). We propose two modes 
for batching: the first is random batching, where data instances are randomly 
assigned to a batch; and the second is cluster batching, where we perform clustering 
on the dataset, and then random batching is conducted within each cluster.


\section{Experiments}
\label{sec:exp}

\subsection{Experimental Setup}
\label{sec:exp-setup}
We use the datasets evaluated in \cite{narayan2022can}. 
We evaluate three LLMs: GPT-3.5-turbo-0301 
(referred to as GPT-3.5), GPT-4-0314 (referred to as GPT-4), and 
GPT-4o-2024-05-13 (referred to as GPT-4o). The temperature parameter for these 
models is set at 0.35. For SM tasks, we use 3 few-shot examples, and 
for other tasks, we use 10. The default batch prompting method is random batching. 
The batch size ranges for GPT-3.5, GPT-4, and GPT-4o are [10, 20], [10, 15], and 
[5, 10], respectively. As baselines, we employ GPT-3 (text-davinci-002) with the best 
settings~\cite{narayan2022can} for all four tasks, and HoloClean~\cite{rekatsinas2017holoclean} 
and HoloDetect~\cite{heidari2019holodetect} for ED, IPM~\cite{mei2021capturing} for DI, 
SMAT~\cite{zhang2021smat} for SM, and Magellan~\cite{konda2016magellan}, Ditto~\cite{li2020deep}, Unicorn/Unicorn ++~\cite{tu2023unicorn}, and Table-GPT~\cite{li2023table} for EM. As these 
baselines have been evaluated in \cite{narayan2022can}, we use these results as a 
reference. Open LLMs like LLaMA are not considered here, as they are generally less competitive than close models~\cite{lmsys}. 

\subsection{Experimental Results}
\label{sec:exp-comparison}

Performance comparisons are presented in Table~\ref{sec:exp-comparison}. GPT-4 surpasses GPT-3 on three out of the four tasks: DI, SM, and EM. For DI and SM, and achieves superior performance than previous methods, particularly for SM. Moreover, GPT-4 emerges as the victor on 4 out of 7 datasets for EM. GPT-3.5 also presents  strong competition, outperforming GPT-3 on DI and SM. GPT-4o is generally on a par with GPT-3.5 on DI and EM, but turns out to be mediocre on ED and SM, showcasing inconsistent performance. Table-GPT, as GPT-3.5 fine-tuned for processing tabular input, roughly exhibits reduced performance from GPT-3.5 on EM. Consequently, we recommend users to either employ larger models or fine-tune its parameters for these tasks, and avoid the model with more HCI focus (i.e., GPT-4o) for the time being. We also observe Ditto, a non-GPT method, excelling on a few datasets. For ED, our performance is not as competitive as the GPT-3 results reported in \cite{narayan2022can}. The prompts used for GPT-3 in \cite{narayan2022can} are not directly applicable for GPT-3.5 and GPT-4. We believe the results of ED warrant further investigation, such as a case-by-case comparison. 

\begin{table} [t]
  \small
  \centering
  \caption{Batch size evaluation, measured on the Adult dataset for ED, using 
  GPT-3.5 without few-shot prompting.}
  \resizebox{\linewidth}{!}{%
  \begin{tabular}{c|c|c|c|c} \hline
    Batch size & F1 score (\%) & Tokens (M) & Cost (\$) & Time (hrs) \\ \hline 
    1          & 44.0 & 4.07 & 8.14 & 4.76 \\
    2          & 45.9 & 2.38 & 4.75 & 2.70 \\
    4          & 45.1 & 1.87 & 3.74 & 2.06 \\
    8          & 45.0 & 1.61 & 3.21 & 1.82 \\
    15         & \textbf{46.3} & \textbf{1.49} & \textbf{2.99} & \textbf{1.60} \\ \hline
  \end{tabular}
  }
  \label{tab:exp-efficiency}
\end{table}


To assess the effectiveness of our prompting strategy, we test GPT-3.5, as it 
is more cost-effective and faster than GPT-4, while delivering notable performance 
in the above evaluations. This makes it a more desirable choice for applications 
dealing with large datasets. The results are reported in 
Table~\ref{tab:exp-ablation-study}. We start with GPT-3.5 prompted with only task 
specification (i.e., without reasoning, as shown in the first line of the prompt 
in Section~\ref{sec:zero-shot}) through zero-shot prompting. The result quality for 
ED and SM is very low, and roughly below 90\% for DI and EM. The inclusion of 
few-shot examples improves all performances, exceeding 50\% for ED and SM and 
reaching approximately 90\% for the others. Batch prompting generally has a slight
negative effect on result quality. With zero-shot reasoning, the performances of ED, 
DI, and SM are further improved, with ED over 90\% and SM over 60\%. However, there 
is little improvement observed for EM, potentially due to GPT-3.5's reasoning 
limitations and the lack of adequate input information for reasoning.

Feature selection proves useful for GPT-4. For instance, for entity 
matching on the Beer dataset without few-shot prompting, the F1 scores before and 
after feature selection are 74.1\% and 90.3\%, respectively. In terms of batch 
prompting, we compare random batching with cluster batching, where data instances 
are clustered using k-means over their Sentence-BERT~\cite{sbert-url} embeddings. For 
entity matching on the Amazon-Google dataset without few-shot prompting, F1 scores 
increase from 45.8\% to 50.6\% when switching from random to cluster batching, 
illustrating the effectiveness of cluster batching.

We explore the impact of batch size and present the results in 
Table~\ref{tab:exp-efficiency}. As batch size augments, there is a significant 
reduction in the number of tokens, dropping from over 4M without batch prompting to 
1.5M with a batch size of 15. Both the cost and processing time follow similar 
trends, decreasing from \$8.14 to \$2.99 and from 4.8 hours to 1.6 hours, respectively. 
Concurrently, the F1 score experiences minor fluctuations, even displaying an increase 
when the batch size is set to 15. This is because GPT-3.5 can identify commonalities in 
questions and generate consistent solutions for all data instances in the batch, 
thereby enhancing overall performance.

\begin{acks}
This work is mainly supported by NEC Corporation and partially supported by JSPS Kakenhi 23K17456, 23K25157, 23K28096, and JST CREST JPMJCR22M2. 
\end{acks}

\balance
\bibliographystyle{abbrv}
\bibliography{references}

\end{document}